\documentclass[sigconf]{acmart}

\AtBeginDocument{%
  \providecommand\BibTeX{{%
    \normalfont B\kern-0.5em{\scshape i\kern-0.25em b}\kern-0.8em\TeX}}}

\usepackage{amsmath}
\usepackage{amssymb}
\usepackage[ruled, linesnumbered, vlined]{algorithm2e}
\usepackage{graphicx}
\usepackage{multirow}
\usepackage{subfigure}
\usepackage{multicol}
\usepackage{threeparttable}
\usepackage{booktabs}
\usepackage{color}

\newtheorem{ETA learning}{Definition}[section]
\newcommand{\F}{\textsl{Fusion RNN}\xspace}

\setcopyright{acmcopyright}
\copyrightyear{2020}
\acmYear{2020}
\acmDOI{x}

\acmConference[DEEPSPATIAL 2020]{1st ACM SIGKDD Workshop on Deep Learning for Spatiotemporal Data, Applications, and Systems}{Augest 24, 2020}{San Diego, California, CA}
\acmPrice{15.00}
\acmISBN{978-1-4503-XXXX-X/18/06}

\begin{document}

\title{Fusion Recurrent Neural Network}

\author{Yiwen Sun}
\affiliation{%
  \institution{Department of Automation, Institute for Artificial Intelligence, Tsinghua University (THUAI)}
  \city{Beijing}
  \country{China}}
\email{syw17@mails.tsinghua.edu.cn}

\author{Yulu Wang}
\affiliation{%
  \institution{Department of Automation, Institute for Artificial Intelligence, Tsinghua University (THUAI)}
  \city{Beijing}
  \country{China}}
\email{wangyulu18@mails.tsinghua.edu.cn}

\author{Kun Fu}
\affiliation{%
  \institution{DiDi AI Labs}
  \city{Beijing}
  \country{China}}
\email{fukunkunfu@didiglobal.com}

\author{Zheng Wang}
\affiliation{%
  \institution{DiDi AI Labs}
  \city{Beijing}
  \country{China}}
\email{wangzhengzwang@didiglobal.com}

\author{Changshui Zhang}
\affiliation{%
  \institution{Department of Automation, Institute for Artificial Intelligence, Tsinghua University (THUAI)}
  \city{Beijing}
  \country{China}}
\email{zcs@mail.tsinghua.edu.cn}

\author{Jieping Ye}
\affiliation{%
  \institution{DiDi AI Labs}
  \city{Beijing}
  \country{China}}
\email{yejieping@didiglobal.com}

\renewcommand{\shortauthors}{Sun, et al.}


\begin{abstract}
  Considering deep sequence learning for practical application, two representative RNNs -- LSTM and GRU may come to mind first. Nevertheless, is there no chance for other RNNs? Will there be a better RNN in the future? In this work, we propose a novel, succinct and promising RNN -- Fusion Recurrent Neural Network (\F).
  \F is composed of Fusion module and Transport module every time step. Fusion module realizes the multi-round fusion of the input and hidden state vector. Transport module which mainly refers to simple recurrent network calculate the hidden state and prepare to pass it to the next time step.
  Furthermore, in order to evaluate \F's sequence feature extraction capability, we choose a representative data mining task for sequence data, estimated time of arrival (ETA) and present a novel model based on \F.  
  We contrast our method and other variants of RNN for ETA under massive vehicle travel data from DiDi Chuxing. 
  The results demonstrate that for ETA, 
  \F is comparable to state-of-the-art LSTM and GRU which are more complicated than \F. 
  
\end{abstract}






\maketitle

\section{Introduction}
\label{sec:Introduction}

How to extract semantic information and analyze pattern of sequence data? 
It is one of the most significant question of sequence learning.
Since 1982, recurrent neural network (RNN) have answered this question step by step though a series of developments\cite{cho2014learning,hochreiter1997lstm,elman1990finding,jordan1997serial,hopfield1982neural}.
Time series data contains complex temporal dependency which is the important semantic information worth extracting.
As a representative deep neural network, RNN is adept in extracting time series features automatically from massive data.
Additionally, RNN is currency and flexible enough to handle sequence data from various fields, such as natural language processing\cite{mikolov2010recurrent}, intelligent transportation system\cite{hofleitner2012learning} and finance\cite{labiad2018short}.

Early RNNs are Simple recurrent networks, Jordan networks\cite{jordan1997serial} and Elman networks\cite{elman1990finding} which capture temporal correlation through recurrent structure. Though alleviating the gradient vanishing and exploding problem, Long Short-Term Memory Network (LSTM)\cite{hochreiter1997lstm} and Gated Recurrent Unit (GRU)\cite{cho2014learning} become state-of-the-art. For LSTM, there are three gates -- the input gate, forget gate, output gate and a memory cell in each inference time step. GRU adopt just two gates and do not adopt the memory cell leading to a simpler structure.

However, the necessity of gates for controling the information flow is worth discussing.
We understand that in essence, gates are to realize the effective interaction between the input and the hidden state, so that the hidden state after training contains the semantic information of the input sequence. 
A more concise and efficient way of information fusion between the input and the hidden state may lead to a next-generation RNN.
Therefore, in this paper, we propose an RNN without any gate and memory cell structure, called Fusion RNN.

The most obvious characteristic of \F is simplicity but effectiveness.
\F has two intuitive and clear modules -- Fusion module and Transport module. For the specific sequential learning task, ETA, a representative spatiotemporal data mining task, is adopted to demonstrate \F's excellent sequential pattern analysis ability.

The main contributions in this work is included:

\begin{itemize}
    
    \item We propose a novel RNN, \F. \F is a succinct and promising RNN in which all gate and memory cell structures are abandoned. Through quantitative calculation, we show that \F is simpler than LSTM and GRU.

    \item For evaluating the feature extraction capability of \F, we present a novel deep learning model based on \F for ETA that is important data mining task. This deep learning model's core sequence feature extraction module of this model can be replaced by any RNN to ensure fair comparison.
   
    \item We train and test our method on the large-scale real-world dataset whose trajectorie number is over 500 million. 
    The experimental results validate that for ETA \F shows competitive sequence feature extraction ability compared with LSTM and GRU.
    
\end{itemize}

We organize this paper's rest part as follows. Section~\ref{sec:RELATED WORK} summarizes the related works of sequence learning as well as recurrent neural network. Section~\ref{sec:METHODOLOGY} introduces the structure of our \F, followed by the deep learning model based on \F. In Section~\ref{sec:EXPERIMENT}, the experimental comparisons on the real-world datasets for ETA are presented to show the superiority of \F. Finally, we conclude this paper and present the future work in Section~\ref{sec:CONCLUSION}.

\section{Related Work}
\label{sec:RELATED WORK}
In this section, we briefly introduce works on sequence learning, recurrent neural network and estimated time of arrival.

\subsection{Sequence Learning}

A large amount of things in life are related to time, which means many living materials are stored in the form of sequence, such as text, speech and video. In order to flexibly handle sequence data of different lengths and extract useful information effectively, sequence learning \cite{cohen1990attention,cui2016continuous,gehring2017convolutional,raganato2017neural,wiseman2016sequence} has become one of the hot spots of machine learning in recent years. It is widely used in recommendation systems \cite{yin2017deepprobe}, speech recognition \cite{chiu2018state,weng2018improving,dong2018speech,cho2018multilingual}, natural language processing \cite{song2019mass,sriram2017cold,chorowski2016towards,wu2017sequence}, 
intelligent transportation system \cite{wang2018learning,fu2020} and other fields. In early works people use support vector machines\cite{cai2003svm}, autoregressive model\cite{lutkepohl2005new}, hidden Markov models\cite{oates2000using} and so on to learn features from sequence. With the increase in the amount of data and the great development of deep learning\cite{lecun2015deep,lecun1995learning,krizhevsky2012imagenet,larochelle2009exploring}, deep neural networks \cite{liu2017survey,wang2018end} have become an effective way to solve sequence learning problems. Among the deep learning methods, recurrent neural network (RNN) is the most effective solution to sequence learning\cite{lipton2015critical}.

\subsection{Recurrent Neural Network}

Recurrent neural network (RNN) is a type of network that takes serial data as input with the ability to selectively pass information across sequence steps. In 1982, Hopfield\cite{hopfield1982neural} first proposed a recurrent neural network, which has pattern recognition capabilities. Elman\cite{elman1990finding} and Jordan \cite{jordan1997serial} improved RNN and their networks are known as “simple recurrent networks” (SRN). Since then researchers continued to make improvements in recurrent neural network \cite{cho2014learning,hochreiter1997lstm,chung2014empirical,yamashita2008emergence,li2018independently,giles1992learning}. The most classical RNN algorithms are two closely related variants of RNN, Long Short-Term Memory Network (LSTM)\cite{hochreiter1997lstm} and Gated Recurrent Unit (GRU) \cite{cho2014learning}. Both LSTM and GRU can carry valuable information for a long sequence, solve the problem of gradient disappearance and gradient explosion in the traditional RNN, and perform well on tasks with long sequences. However, several gates have been used in both GRU and LSTM structures, which greatly improve the complexity of the model. Our model which simply carry interactive operations between input vectors and hidden states doesn't add additional gates into a cell. We can achieve a better performance than LSTM and GRU in the sequential learning task with a simpler structure.

\subsection{Estimated time of arrival}
Estimated time of arrival (ETA) is one of the difficult problems in sequence learning, which refers to the process of estimating the travel time when given departure, destination and the corresponding route. It is a challenging problem in the field of intelligent transportation system. There are two category methods to solve the problem, route-based method and data-driven method. The route-based method focus on estimating the travel time in each individual link which is split from the original whole route. Traditional machine learning methods such as dynamic Bayesian network  \cite{hofleitner2012learning},  gradient boosted regression tree \cite{zhang2016urban}, least-square minimization \cite{zhan2013urban} and pattern matching\cite{chen2013dynamic} are typical approaches to capture the temporal-spatial features in the route-based problem. However, a drawback of this method is the accumulation of local errors brought by the split operation.

The data-driven method is popular both in academia and industry. As the advance of deep learning\cite{lecun2015deep,larochelle2009exploring}, this category method has shifted from early traditional machine learning methods such as TEMP\cite{wang2019simple} and time-dependent landmark graph\cite{yuan2011t} to deep learning methods. Deeptravel\cite{zhang2018deeptravel} converts the travel trajectory into a grid sequence, and learns the time interval between the origin and the destination to the intermediate point simultaneously through BiLSTM. DeepTTE\cite{wang2018will} adopts propsed geo-convolution and standard LSTM to estimate the travel time through raw GPS points. WDR\cite{wang2018learning} proposes a joint model of wide linear model, deep neural network and recurrent neural network, which can effectively capture the dense features, high-dimensional sparse features and local features of the road sequence. Most data-driven methods use LSTM for time series’ learning but the LSTM model is slow on large-scale data sets. So it’s vital to further simplify the LSTM model and retain its good performance in ETA. 
\section{Method}
\label{sec:METHODOLOGY}
In this section, we introduce the architecture of \textit{Fusion RNN} and its application in estimated time of arrival (ETA).
\begin{figure*}[htbp]
   \centering
   \includegraphics[width=0.85\textwidth]{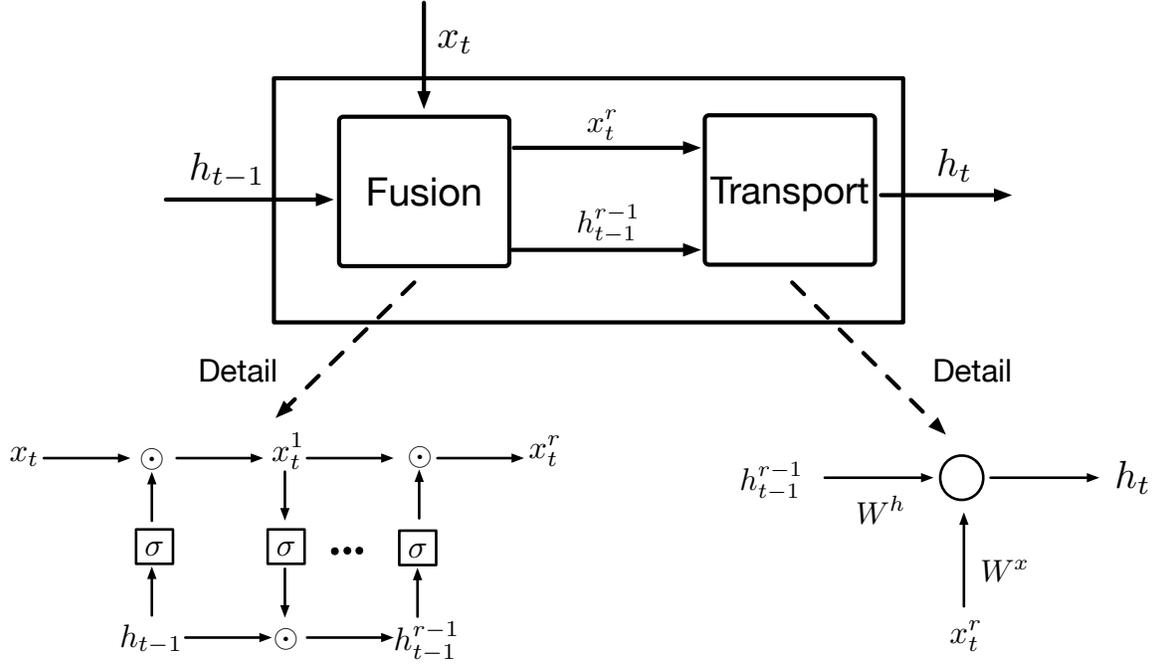}
   \caption{The architecture of Fusion RNN. Information in the input vector and the hidden state vector interact in the Fusion module which is shown in the bottom left part of the figure. Here the number of iterations $r$ is odd. The Transport module is a basic recurrent unit which is shown in the bottom right part of the figure.}
   \label{fig:fusion}
\end{figure*}

\subsection{Fusion RNN}

Gated recurrent neural network\cite{cho2014learning} has achieved promising results in many sequence learning tasks. It uses several gating units to control the transfer process of information in the network. LSTM \cite{hochreiter1997lstm} is one of the most widely used RNN algorithms because of its excellent performance especially on long sequence tasks. However, LSTM adds three gated units, an input gate, an output gate and a forget gate, on the basis of simple recurrent networks. The gates greatly increased the amount of model parameters and the computational complexity.
GRU's structure is like LSTM  but has fewer parameters than LSTM, as it lacks an output gate. GRU's performance on certain tasks of polyphonic music modeling and speech signal modeling was found to be similar to that of LSTM. GRU have been shown to perform better performance than LSTM on some smaller datasets\cite{chung2014empirical}. However, GRU still has two gates, the reset gate and the update gate. Is gated unit the only option for RNN to achieve good performance? Can we use a recurrent neural network without gated units to achieve similar performance in sequence learning tasks? To 
solve this problem, our strategy is to directly make changes on the architecture of simple recurrent networks. We propose a novel recurrent neural network, \textit{Fusion RNN}, without any gated unit. We achieve a similar performance with LSTM and GRU on sequence learning tasks through a very simple interactive operation structure. We will discuss \textit{Fusion RNN} in detail next.

 As shown in Figure \ref{fig:fusion}, the large black box in the upper part of the figure represents one time step, and the hidden state parameter $h_t$ of the next time step should be calculated by the input $x_t$ and the hidden state $h_{t-1}$ of the previous time step. As for how to calculate $h_t$ and how to control the information flow, there are different designs in different recurrent neural networks. For example, Elman network uses a simple recurrent unit while LSTM uses three gates in each time step. We think that the most important thing is the fusion of input information and the hidden state information and the calculation of h. So we designed two modules in our \F, Fusion module and Transport module. The Fusion module is a process of multi-round interactive fusion of input vector $x_t$ and the hidden state vector $h_{t-1}$. The purpose of the fusion process is make the final input vector ${x_t}^r$ contains information of ${h_{t-1}}$, and the final hidden state $h^{r-1}_{t-1}$ contains information of ${x_{t}}$. We refer to \cite{melis2019mogrifier} for the mogrifier operation of input vector $x_t$ and the hidden state parameter $h_{t-1}$ between each time step. We don't add the hyperparameter 2 to our Fusion module's formula so that the model is easier to converge. This makes our proposed structure more essential and significant. Full fusion of features is the foundation of the Transport module. The Transport module is a recurrent unit in which the current hidden state $h_t$ is calculated. We concatenate the input $x_t$ and the previous hidden state $h_{t-1}$ together, and use a feed forward network to extract features based on on the simple recurrent network, Elman network. Here we select tanh function as the activation function. Then the current hidden state $h_t$ is passed to the next time step.
 \begin{figure*}[htbp]
   \centering
   \includegraphics[width=0.6\textwidth]{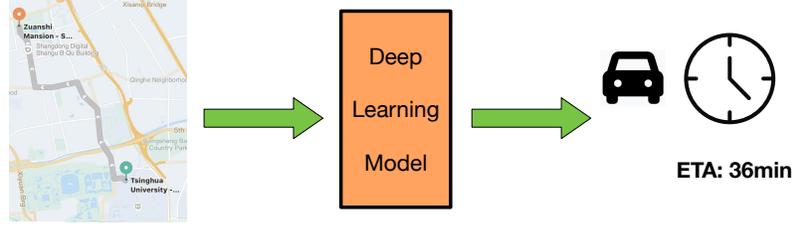}
   \caption{The conceptual sketch of deep learning based ETA. ETA is the travel time along the given path for a specific vehicle. A deep learning model is adopted for automatic extraction of semantic features from mass data.}
   \label{fig:ETAapp}
\end{figure*}

In order to intuitively reflect the simplicity of our model, we compare the parameters of Fusion RNN with LSTM and GRU models. We can see from table\ref{tab_para} that our model have fewer parameters than GRU and LSTM. \textit{Fusion RNN}'s amount of parameters is $n^2+3mn+n$, $2n^2+2n$ less parameters than GRU and $3n^2+mn+3n$ less paremeters than LSTM when $m\not= n$. \textit{Fusion RNN}'s amount of parameters is $4n^2+n$, $2n^2+2n$ less parameters than GRU and $4n^2+3n$ less paremeters than LSTM when $m=n$. 
The interactive operation in Fusion module is calculated by Formula \ref{formula1} and Formula \ref{formula2}, the recurrent unit in Transport module is calculated by Formula \ref{formula3}:
\begin{equation}
\label{formula1}
{x}_{t}^{i}=\sigma\left({F}^{ {x}} {h}_{t-1}^{i-1}\right) \odot {x}_{t}^{i-2}, \qquad \text { for odd } i \in[1 \ldots r] \\
\end{equation}
\begin{equation}
\label{formula2}
{h}_{t-1}^{i}=\sigma\left({F}^{{h}} {x}_{t}^{i-1}\right) \odot {h}_{t-1}^{i-2},\qquad  \text { for even } i \in[1 \ldots r] 
\end{equation}
\begin{equation}
\label{formula3}
h_{t}=\tanh\left(W^{x} x_{t}+W^{h} h_{t-1}+b\right)
\end{equation}
where $x_t$ is the input vector, $h_{t-1}$ is the hidden layer vector, $y_t$ is the output vector, $W^x,W^h$ is weight, $b$ is bias, $r$ is the number of iterations, $F^{*}$ are learned parameters, $\tanh$ is the tanh function, and $\sigma$ is the activation functions. The activation functions commonly used in RNN are sigmoid function and tanh function. Here we use $\tanh$ function as activation function in Fusion module. 
\begin{table}[!t]
\caption{The amount of parameters in different RNN models.}
\label{tab_para}
\tabcolsep 12pt 
\begin{threeparttable}
\begin{tabular*}{0.45\textwidth}{ccc}
\toprule
                 &$m\not =n$ &  $m=n$\\
\midrule
 LSTM    &   $4mn+4n^2+4n$ &  $8n^2+4n$ \\
 GRU  &   $3mn+3n^2+3n$   &   $6n^2+3n$ \\
 Fusion RNN        &   $n^2+3mn+n$  &  $4n^2+n$ \\
\bottomrule
\end{tabular*}
\begin{tablenotes}
    \footnotesize
     \item[1] $m$ is the input size and $n$ is the hidden size.
\end{tablenotes}
\end{threeparttable}
\end{table}

Here $m$ is the input size and $n$ is the hidden size and usually $m<n$. The results shows that our \textit{Fusion RNN}'s training time is shorter and the model size is smaller than other gate-based recurrent neural networks which means that our model is more convenient to application.

\begin{table}[!t]
\caption{Multiplication times of one recurrent cell.}
\label{tab_multi}
\begin{threeparttable}
\begin{tabular*}{0.45\textwidth}{ccc}
\toprule
                 & $m\not=n$ &$m=n$\\
\midrule
 LSTM    &   $4mn+4n^2+3n$ & $8n^2+3n$ \\
 GRU  &   $3mn+3n^2+3n$   & $6n^2+3n$  \\
 Fusion RNN & $n^2+(1+r)mn+([r/2])n$& $(r+2)n^2+rn$\\
 & $+([r/2]+1)(m)$ & \\
\bottomrule
\end{tabular*}
\begin{tablenotes}
    \footnotesize
     \item[1] $[.]$ is the floor function, $r$ is the number of interaction iterations in Fusion module, $m$ is the input size and $n$ is the hidden size. The length of sequence $l$ is suppose to be 1. Otherwise, all items in this table should be multiplied by $l$.
\end{tablenotes}
\end{threeparttable}
\end{table}

 We have iterative interactive operations in \textit{Fusion RNN}. Considering the effect of iterations on the model's inference speed, we compared the multiplications times of one recurrent cell in different RNN algorithms. The relusts is shown in Table \ref{tab_multi}. Here the length of sequence $l$ is suppose to be 1. Otherwise, all items should be multiplied by $l$.\textit{Fusion RNN}'s amount of parameters is $n^2+(1+r)mn+([r/2])n+([r/2]+1)(m)$ when $m\not=n$ and $(r+2)n^2+rn$ when $m=n$. The multiplication times of \textit{Fusion RNN} is depend on the number of iterations $r$, which is usually set between 1 and 5. From the table \ref{tab_multi} we can learn that when $r$ is less than 4, Fusion RNN has fewer multiplication times than GRU and when $r$ is less than 6, Fusion RNN has fewer multiplication times than LSTM. The relult shows that in general, \textit{Fusion RNN} is more concise than LSTM which means our model's inference speed is faster than LSTM. Experiments in Section \ref{sec:EXPERIMENT} shows that our \textit{Fusion RNN} performs well in sequence learning. Our \textit{Fusion RNN} outperforms GRU and performs a little better than LSTM. The detail will be disscussed in Section \ref{sec:EXPERIMENT}. 
 
 Our \textit{Fusion RNN} only uses three formulas. Its amount of parameters and computational complexity are less than LSTM. That is, we successfully propose a simpler recurrent neural network whose performance is close to or even better than LSTM without using any gated unit.

\subsection{FusionRNN-ETA}
\begin{figure*}[htbp]
   \centering
   \includegraphics[width=1\textwidth]{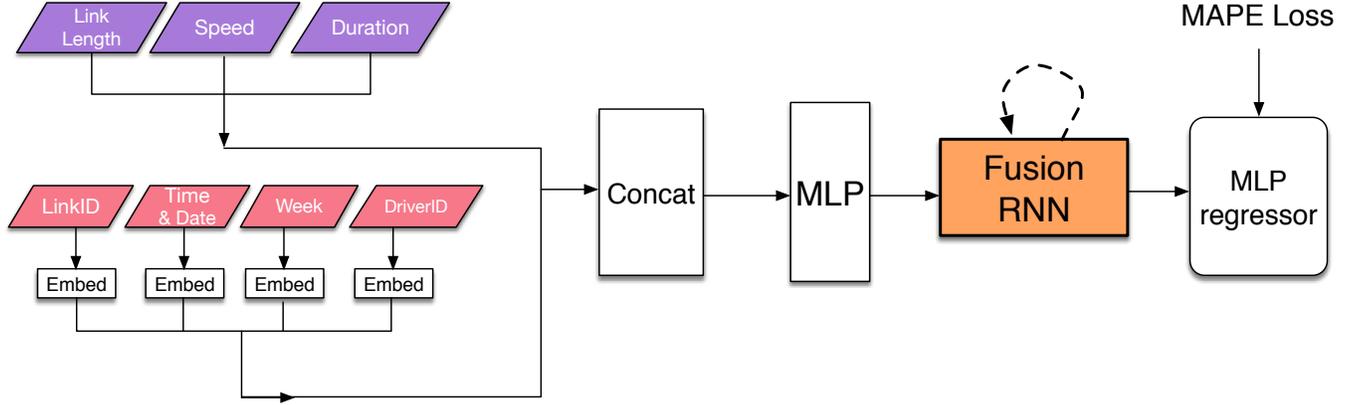}
   \caption{The architecture of FusionRNN-ETA. Spartial features are concated with sequetial features after embedding layer. After the MLP module, features enter the feature extractor Fusion RNN. Finally, we perform regression on the extracted features.}

   \label{fig:fusionETA}
\end{figure*}
Estimated time of arrival (ETA) is the process of estimating the arrival time based on departure, destination and trajectory information. Figure \ref{fig:ETAapp} shows an example of ETA and one of the practical application of ETA. It is one of the most basic and essential tasks in the intelligent transportation system and one of the representative tasks of sequence learning based on spatio-temporal data.
We test the performance of \textit{Fusion RNN} on ETA problem and propose a FusionRNN-ETA model. Wang et al. modeled ETA as a regression learning problem. We then give the definition of ETA learning problem which is essentially a regression task:
\begin{definition}[\textbf{ETA learning}]\label{defETA}
For a track order from massive data $D=\left\{s_{i}, e_{i}, d_{i}, \boldsymbol{p}_{i}\right\}_{i=1}^{N}$,  $s_{i}$ is i-th trak departure time, $e_{i}$ is i-th trak arrival time, $d_i$ is the driver ID, $\boldsymbol{p}_{i}$ is the road segment sequence set of the order. $\boldsymbol{p}_{i}$ is represented as $\boldsymbol{p}_{i}=\left\{l_{i 1}, l_{i 2}, \cdots, l_{i T_{i}}\right\}$, where $l_{ij}$ is the j-th link of i-th trajectory, and $T_i$ is the length of the link sequence set. $N$ is the total sample number of the dataset. We computed the ground truth travel time by $y_{i}=e_{i}-s_{i}$. Our objective is to train a machine learning model which could estimate the travel time $ y_{i}^{\prime}$ when the departure time, the driver ID and the route are given.
\end{definition}
\subsubsection{Data Preprocessing}
We first construct a rich feature set from the raw information of trips. The original data in the dataset of ETA tasks is made up of the departure's and destination's latitude and longitude, the time stamp, driver ID, and GPS points of the trajectory. We convert the time stamp to date, day of the week and specific departure time. We divide the day into 288 5-minutes time slices. Then we carry map matching \cite{newson2009hidden} algorithm on the GPS trajectory to obtain a set of road links corresponding to the set of GPS points.
We use historical trajectories in adjacent previous time to estimate the road conditions. After the process, the features can be categorized into two kind of types: 

(1) The non-sequential features which are irrelative to the travel path, such as time and driver ID. 

(2) The sequential features extracted from the link sequence set $\boldsymbol{p}_{i}$. For j-th link $l_{ij}$ in $\boldsymbol{p}_{i}$, we denote its feature vector as $x_{ij}$ and get a feature matrix $X_i$ for the i-th trip. The features in $x_{ij}$ is the basic information of the selected link, e.g. length and speed.

\subsubsection{Overall framework}
We desire to adopt our Fusion RNN to ETA problem. We choose the state-of-the-art Wide-Deep-Recurrent (WDR) model \cite{wang2018learning} as our basic model to solve the problem. The three components of WDR model includes: 
(1) A wide module uses a second-order cross product and affine transformation of the non-sequential feature$z_i$ to memorizes the historical patterns;
(2) A deep module constructs embedding table\cite{bengio2003neural} for the sparse discrete features in sequence. The embedding vectors are concatenated, then they are put into a Multi-Layer Perceptron (MLP), which is a stacked fully-connected layers with ReLU\cite{krizhevsky2012imagenet} activation function;
(3) A recurrent module integrates the sequential features together and inputs these features into a fine-grained RNN-based model which can capture the spartial and temporal dependency between links. 

We simplify and further improve the structure of WDR and develop a novel ETA model FusionRNN-ETA based on our \F which have interactive operation between the input vector and the hidden state parameter in each time step. We remove the wide module of WDR. The principles of \F have been explained in detail in Section 3.1. Next, we will introduce the overview of our model and describe all components in detail.A description of the overall the architecture of FusionRNN-ETA is given in Figure \ref{fig:fusionETA}. We remove the wide module in WDR and directly use deep module to extract sequential and non-sequential features. We use embeddings on link ID and non-sequential features. The embedding vector is an important component of these features. For example, for a link with ID $l_{ij}$, we look up an embedding table\cite{bengio2003neural} $\boldsymbol{E}_{L} \in \mathbb{R}^{20 \times M}$, and use its $l_{ij}$-th column $\boldsymbol{E}_{L}\left(:, l_{i j}\right)$ as a distributional representation for the link. We perform the same embedding operation on the 
non-sequential features.

Next we concatenate the embedding features with the sequential features. Then we put the concatenated features into a MLP which is a stack of fully-connected layers with RELU activation function. The output features after MLP are feed into a recurrent module. We use \F as the recurrent neural network algorithm in the recurrent module. The travel time prediction is given by a regressor, which is also a MLP, based on the outputs of \F. 
The hidden state sizes in \F and the regressor MLP are all set to 128. The hidden state of \F are initialized as zeros.

The parameters of the FusionRNN-ETA are trained by optimizing the Mean Absolute Percentage Error (MAPE) loss functions:
\begin{equation}
\mathrm{MAPE}=\frac{1}{N} \sum_{j=1}^{N} \frac{\left|y_{j}-y_{j}^{\prime}\right|}{y_{j}}
\end{equation}
where $ y_{j}^{\prime}$ is the predicted travel time, $y_j$ is the ground truth travel time.

\section{Experiment}
\label{sec:EXPERIMENT}
\subsection{Dataset}
The dataset used in our experiment is the massive real-world floating-car trajectory dataset provided by the Didi Chuxing platform. It contains the trajectory information of Beijing taxi drivers for more than 4 months in 2018, so we call this dataset Beijing 2018. The dataset is huge and the numbers of unique links are very large. These road-segments contain a variety of complicated types, for instance, local streets and urban freeways. 
Before the experiment, we filtered the anomalous data whose travel time less than 60s and  travel speed greater than 120km/h. We divided this data set into a training set (the first 16 weeks of data), a validation set (the middle 2 weeks of data) Test set (data for the last 2 weeks).

\subsection{Compared Methods}
We compare our FusionRNN-ETA with 6 route-based, linear-based, and RNN-based models on the massive real-world datasets. We select route-ETA as the representation of traditional method. It has vety fast inference speed but its accuracy is often far from satisfactory compared with deep learning methods. We choose an improved version of WDR which is illustrated in Section 3.2 as the representative of deep learning methods. We apply different RNN algorithms on the recurrent module of the deep model or use other deep models such as FFN and resnet to replace the recurrent module.

(1) Route-ETA: a representative method for non-deep learning methods. It splits a trajectory into many links. The travel time $t_i$ in the i-th link in this trajectory is calculated by dividing the link’s length by the estimated speed in this link. The delay time $c_j$ in the j-th intersection is provided by a real-time traffic monitoring system. The final estimated time of arrival is the sum of all links’ travel time and all intersections’ delay time.

(2) FFN: an artificial neural network where the connections between each nodes don’t form a cycle. Here we use a Multi-Layer Perceptron network for comparision.

(3) Resnet\cite{he2016deep}: a deep convolutional neural network with residual structure. It explicitly reformulate the layers as learning residual functions with reference to the layer inputs \cite{he2016deep}. Here we use a shallow simplified version to extract features.

(4) RNN: an artificial neural network which takes sequence data as input and recurs in the direction of sequence. The connections between nodes form a directed graph along the temporal sequence. Here we use Elman network \cite{elman1990finding} as RNN algorithms. 

(5) GRU \cite{cho2014learning}: a specific recurrent neural network with a gating mechanism. GRU has two gates, the reset gate and the update gate.

(6) LSTM \cite{hochreiter1997lstm}: a specific recurrent neural network with three gates, an input gate, a forget and an output gate on the basis of simple recurrent networks.

\begin{table}[!t]
\caption {The results of different sequence pattern analysis methods}
\label{tab_loss}
\footnotesize
\begin{tabular*}{0.4\textwidth}{cccc}
\toprule
               & MAPE(\% )& MAE(sec) &  RMSE(sec)\\
\midrule
 Route-ETA & 25.010 & 69.008 & 106.966 \\
 FFN         &   21.106        &  57.797    &   93.588\\
 Resnet       &   21.015      &  57.064 &     92.241 \\
 RNN & 19.677 &55.284 & 90.836\\
 GRU     &   19.673        &  55.372  &   90.801\\
 LSTM     &   19.598     & 55.227  &   90.480\\
 \textbf{Fusion RNN (ours)}      &  \textbf{19.579}      &  \textbf{55.019}&     \textbf{90.289}  \\
\bottomrule
\end{tabular*}
\end{table}
\subsection{Experimental Settings}

In our experiment, all models are written in PyTorch and are trained and evaluated on a single NVIDIA Tesla P40 GPU. The number of iterations of the deep learning-based method is 3.5 million, and the batch size is 256. FusionRNN-ETA's number of iteration is $ r $ ($ r \ in [1,5] $). We use the method of Back Propagation (BP) to train the deep learning-based methods, and the loss function is the MAPE loss. We choose Adam as the optimizer due to its good performance. The initial learning rate is 0.0002.

\subsection{Evaluation Metrics}
\begin{figure}[]
\includegraphics[width=0.8\linewidth]{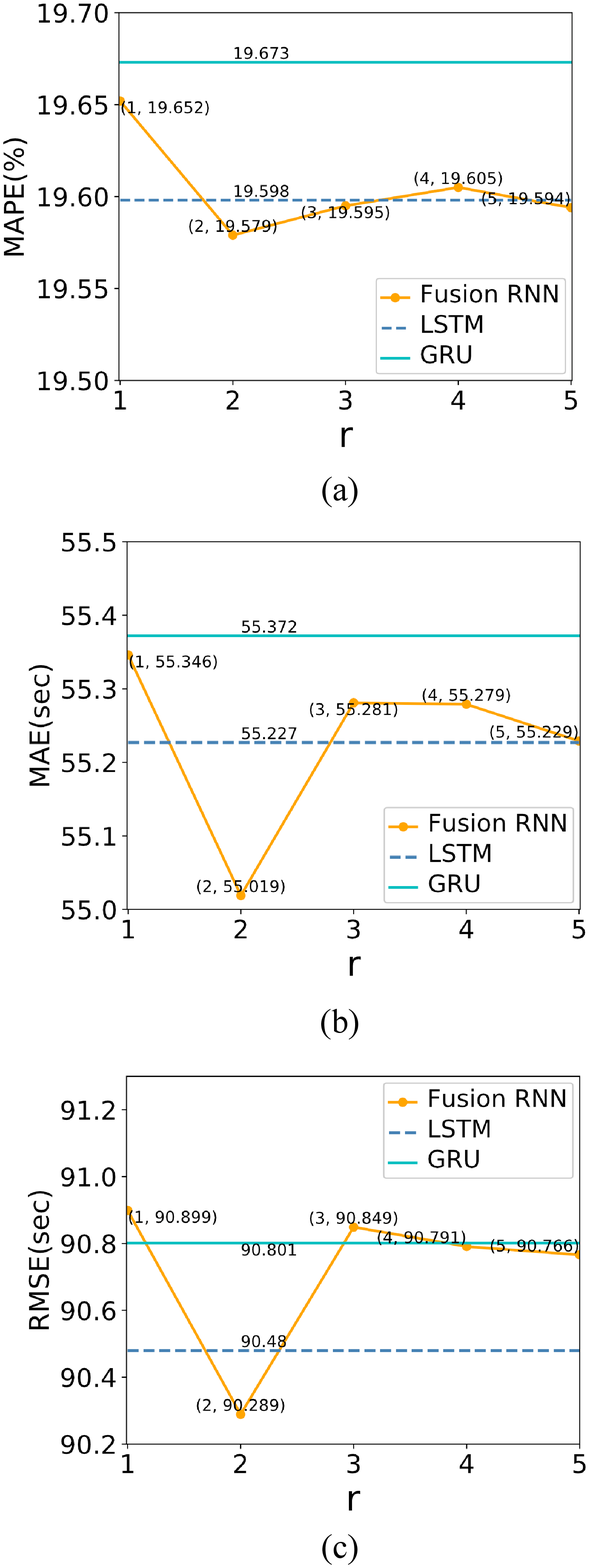}
\caption{FusionRNN-ETA's performance under different number of iterations.
(a) Comparison in terms of MAE on Beijing 2018 dataset. 
(b) Comparison in terms of MAPE on Beijing 2018 dataset. 
(c) Comparison in terms of MSE on Beijing 2018 dataset.}
\label{fig:loss}
\end{figure}
To evaluate and compare the performance of FusionRNN-ETA and other baselines, we use three popular criterions, Mean Absolute Percentage Error (MAPE), Mean Absolute Error (MAE) and Rooted Mean Square Error (RMSE):
\begin{equation}
\mathrm{MAE}=\frac{1}{N} \sum_{i=1}^{N}\left|y_{i}-y_{i}^{\prime}\right|
\end{equation}
\begin{equation}
\mathrm{RMSE}=\sqrt{\frac{1}{N} \sum_{i=1}^{N}\left(y_{i}-y_{i}^{\prime}\right)^{2}}
\end{equation}
where $ y_{j}^{\prime}$ is the predicted travel time, $y_j$ is the ground truth travel time, and $N$ is the number of samples.
The calculation process of MAPE is shown in Section 3.1.

\subsection{Comparision and Analysis of Results}

As seen in the Table \ref{tab_loss}, our \F outperforms all competitors in terms of all metrics on Beijing 2018 dataset. The detailed experimental results and analysis are as follows.

(1) The representative non-deep learning method, route-ETA obviously performs worse than other deep learning methods. Route-ETA is a rule-based method baseline. This result shows that the data-driven method is more effective than the artificial method in modeling complex transportation system given massive spatio-temporal data.

(2) LSTM and GRU performs better than RNN which is specifically Elman Network. Among the existing variants of RNN, LSTM performs best. 
This is because the introduction of gating mechanism really improves the mining ability for sequence data.

(3) Our \F performs best for ETA and outperforms LSTM by $0.09 \%$ in terms of MAPE loss, $0.38 \%$ in terms of MAE loss and $0.21 \%$ in terms of RMSE loss. 

Our Fusion RNN can achieve better performance with a simpler structure than LSTM suggesting that \F is a promising universal sequential feature extractor.
\subsection{Influence of Hyper-parameter}

We explore the impact of the number of iterations in Fusion module on the model's performance, and the results are shown in Fig. As illustrated by the graphs, the following results can be summarized. 

(1) No matter how the parameter $r$ changes from 1 to 5, \F is competitive compared with LSTM and GRU. This demonstrates that \F is robust according to $r$. Furthermore, when $r = 1\  or\  2$, \F is pretty simpler than LSTM, yet, has similar prediction effect.

(2) The superiority of \F compared with LSTM and GRU is obvious when $r = 2$. This shows that the proper number of fusion rounds of input and hidden state vector is of importance for sequence semantic analysis.


\section{Conclusion and Future Work}
\label{sec:CONCLUSION}
In this paper, we propose a novel RNN -- Fusion Recurrent Neural Network (\F). \F is a general, succinct and effective feature extractor for sequence data. 
\F is simpler than state-of-the art and popular LSTM and GRU.
We conduct sufficient experiments on the large-scale real-world dataset for one representative data mining task, ETA.
The evaluations show that \F has competitive sequence semantic information extraction ability compared with LSTM and GRU.
Evaluating and tapping the potential of \F in other sequence machine learning domain, such as natural language processing and speech processing is our future work. Maybe this paper is just a small step, nonetheless, a better RNN is of great importance for the future of sequence learning.


\bibliographystyle{ACM-Reference-Format}
\bibliography{sample-base}

\appendix

\end{document}